\documentclass[
]{ceurart}

\usepackage{listings}
\usepackage[T1]{fontenc} 

\usepackage{latexsym}
\usepackage{graphicx}
\usepackage{tabularx} 
\usepackage{tcolorbox} 

\begin{document}

\copyrightyear{2026}
\copyrightclause{Copyright for this paper by its authors.
  Use permitted under Creative Commons License Attribution 4.0
  International (CC BY 4.0).}

\conference{}

\title{Validating DBpedia Triple Sets for Natural Language Generation}

\author[1]{Mark Andrade}[%
orcid=0009-0000-0575-7377,
email=mark.andrade@adaptcentre.ie
]
\cormark[1]
\address[1]{ADAPT Centre, Dublin City University, Ireland}
\address[2]{University of Bordeaux, CNRS, LaBRI-UMR5800, France}

\author[1,2]{Simon Mille}[%
orcid=0000-0002-8852-2764,
email=sfmille@gmail.com,
]
\cormark[1]

\author[1]{Anya Belz}[%
orcid=0000-0002-0552-8096,
email=anya.belz@adaptcentre.ie ,
]

\author[1]{Brian Davis}[%
orcid=0000-0002-5759-2655,
email=brian.davis@adaptcentre.ie ,
]

\cortext[1]{Corresponding author.}

\begin{abstract}
We present a study of the quality of individual DBpedia triples from the perspective of Natural Language Generation, and propose and evaluate an approach for collecting entity-specific triple sets that filters out questionable triples while minimizing the loss of correct ones. We show in an evaluation against manually annotated data that with validation rules, it is possible to reach 98$\%$ precision in triple selection, and with improvements to a few Property definitions, it is possible to improve recall by 40$\%$ without harming precision.
\end{abstract}

\begin{keywords}
  Natural Language Generation \sep
  DBpedia triples \sep
  Dataset \sep
  Validation
\end{keywords}

\maketitle

\section{Introduction}
\label{sec:intro}

Knowledge graphs such as DBpedia~\cite{lehmann2015dbpedia} are primarily used for Information Extraction purposes: while the correctness of facts is important, their coverage is usually more relevant. For data-to-text Natural Language Generation (NLG), which consists in converting a series of non-linguistic facts into a well-formed text~\cite{reiter2000building} in a given language, inputs should contain only correct information: when considering a series of facts about an entity, it is more important to generate meaningful and factually correct texts than to generate exhaustive at the risk of including nonsensical or factually incorrect information. 

In the present paper, we present a study of the quality of the current state of individual DBpedia triples from the perspective of Natural Language Generation, and propose and evaluate an approach to filter out questionable triples while limiting the amount of correct triples lost in the way. Existing work reports on how to assess the quality of DBpedia as a whole, including aspects such as consistency and relevancy of the facts found on DBpedia~\cite{zaveri2013user}, where facts are under the form of triples \texttt{Subject} $\|$ \texttt{Property} $\|$ \texttt{Object}, e.g. \texttt{Barack$\_$Obama} $\|$ \texttt{birthYear} $\|$ \texttt{1961}. In this paper, our scope is reduced to the following: (i) We only look at triples in isolation: we are particularly interested in whether or not a specific triple can be included in the input of an NLG system, that is, if this triple conveys ``valid'' information. We not look for inconsistencies between triples, or for the coverage that a triple set has of the knowledge about an entity. (ii) We only assess the semantic correctness of each triple against the DBpedia ontology model: we are not interested in, e.g., the pitfalls of the ontological model itself, which is a highly specialised tasks that fall out of our area of expertise. (iii) We only use information that can be found on existing knowledge repositories: we avoid fixing errors by prediction, so as to reduce the risk of introducing new errors in the process.

In the remainder of the paper, we:
\begin{itemize}
    \item Present our quality assessment of individual DBpedia triples: we collect $\sim$600,000 unique triples for $\sim$11,000 entities, put a series of checks in place to validate triples, and find that a small set of Properties is responsible for a large proportion of absence of validity (Section~\ref{sec:qualityAssessment}).
    \item Use our validation strategy to filter out suspicious triples, propose lightweight modifications to recover triples that were actually good but ended up being filtered, and assess the precision and recall of our approach, showing that we can get up to 0.98 precision for the triple selection, while maintaining satisfactory level of recall at around 0.75 (Section~\ref{sec:datasetTriples}). 
    \item Discuss some related work (Section~\ref{sec:SoA}) and concluding remarks (Section~\ref{sec:futureWork}). 
\end{itemize}

All the code for collecting, validating and fixing triples, as well as the code to compute the assessments provided in the paper can be found at \url{https://github.com/andradeM17/DBpedia}.

\section{Quality assessment of DBpedia triples}
\label{sec:qualityAssessment}  
In this section, we introduce a strategy for validating triples (Section~\ref{subsec:method-validation}) and apply it on a large scale to determine the extent to which DBpedia triples contain validated knowledge (Section~\ref{subsec:method-issues}) and identify the main causes of invalidity (Section~\ref{subsec:method-diagnosis}). First, in Section~\ref{subsec:entity-selection}, we describe the process for selecting the entities used to query triples, while in Section~\ref{subsec:triple-selection}, we outline the method for collecting the triples.

\subsection{Entity selection}
\label{subsec:entity-selection}
Three different lists of entities were created for our experiments: two lists of ``popular'' entities, and one list of random entities. The popular entities have more triples than other entities on DBpedia, and these triples come from higher-quality Wikipedia entries, and are more likely to have been manually validated. The first two lists were created using the Wikimedia Foundation ``List of articles every Wikipedia should have''. The first set of entities (\textbf{Top 1,000}) was taken from the main list, which consists of 1,000 articles~\citep{meta_wiki_articles_list}, while the second set (\textbf{Top 10,000}) was compiled from the expanded version, which consists of approximately 10,000 articles \citep{meta_wiki_expanded_articles_list}. This number was 9,995 in December 2025 and 9,999 in March 2026, when the entities used were extracted. In July 2026 it did contain exactly 10,000 articles.\footnote{Due to some discrepancies in the formatting between Wikipedia and DBpedia, some page names had to be manually checked and changed.} Both of these lists categorise their pages according to the thematic content of the articles, for example the Top 10,000 includes \textit{China} under Geography, the \textit{Odyssey} under Language and Literature, and \textit{Turkey (bird)} under Biology and health sciences. These classifications are then used for a category-level analysis in Section~\ref{subsec:method-issues}.

The third list of 1,000 entities (\textbf{Random 1,000}) was selected randomly, through an API call to Wikipedia for a set of entities, excluding all pages with a colon (``:'') in their title, as these would mostly be Category pages, Template pages, and Special pages (for example, "Category:Political terminology of the United States"). Selecting random entities allows us to get a more real-life picture of the general quantity and quality of triples on DBpedia, since the Top 1,000 and the Top 10,000 represent only a very small subset of all entities. 
Unlike the Top K entities, the random entities do not come with a category. In order to have the same categories available for our analysis in Section~\ref{subsec:method-issues}, the random entities were classified according to the top-level categories from the Top 1,000 dataset, by prompting Claude Sonnet 4. In order to assess the classification ability of Sonnet 4, a separate set of 100 DBpedia entities were assigned one of the \textbf{Section} labels from~\citep{meta_wiki_articles_list}. 
Two human annotators (authors) evaluated the classification by comparing the random entities to the existing entities under the section labels of~\citep{meta_wiki_articles_list} and marking each of the annotations as "Correct" or "Incorrect". The annotators achieved 0.96 and 0.95 observed agreement with Sonnet 4 for Section-level, and 0.97 with one another, with a Cohen's $\kappa$ of 0.65 \citep{cohen1960kappa}, see Table~\ref{tab:claude-classification}. This show \textit{substantial} agreement, in terms of the agreement strength divisions of \citet{landis1977measurement}. We also evaluated more fine-grained labels although they are not used in our analysis, i.e. \textbf{Subsection} and \textbf{Subsubsection}, which respectively show \textit{moderate} and \textit{fair} agreement.

\begin{table*}[ht]
\caption{Observed agreement between Claude and human annotators for Section, Subsection and Subsubsection classifications. Only Section labels are used i our analysis.}
\centering
\begin{tabularx}{\textwidth}{|X|c|c|c|}
\hline
\textbf{Author Comparison} & \textbf{Section} & \textbf{Subsection} & \textbf{Subsubsection} \\ \hline
Human A and Claude           & 96$\%$             & 72$\%$                & 67$\%$                   \\ \hline
Human B and Claude            & 95$\%$             & 81$\%$                & 96$\%$                   \\ \hline
Human A and Human B           & 97$\%$             & 83$\%$                & 69$\%$                   \\ \hline \hline
Plain Cohen's kappa between humans          & 0.65             & 0.53                & 0.21                   \\ \hline
\end{tabularx}
\label{tab:claude-classification}
\end{table*}

\subsection{Triple Selection}
\label{subsec:triple-selection}

DBpedia contains both DBpedia raw Properties (\texttt{dbp:}) and DBpedia ontology Properties (\texttt{dbo:}). \texttt{dbp:} Properties represent the raw data as extracted from Wikipedia. Properties that are adjusted to match the DBpedia schema are ontology Properties. For example the triple \textit{dbr:China $\|$ dbp:percentWater $\|$ 2.800000} adjusted to the schema becomes \textit{dbr:China $\|$ dbo:percentageOfAreaWater $\|$ 2.800000 (xsd:float)}. This \texttt{dbo:} triple consists of a Subject (a DBpedia resource, or \texttt{dbr:}), a Property, and an Object (an XML Schema Definition float, or \texttt{xsd:float}. Objects may also be \texttt{dbr:} values.

\texttt{dbo:} triples where the selected entities are either in the Subject or Object position were retrieved using the SPARQL queries in the box below. The \texttt{dbo:} Properties were selected as expected types for their Subject and Object are usually available and can be used to check that the actual type of the Subject and/or Object matches, unlike their \texttt{dbp:} equivalents. In most cases, there are more triples with the selected entities as the Object in triples: the entities in the Top 1,000 list had an average of 8 triples in which they are in the Subject position, and 112 triples in which they are in the Object position.

\begin{tcolorbox}[
    width=\columnwidth,
    title=SPARQL Queries used for triple extraction,
    label=figure:triple_extraction_query
]
SELECT ?property ?value WHERE \{

    \hspace{1cm}<http://dbpedia.org/resource/\{entity\}> ?property ?value .

    \hspace{1cm}FILTER(STRSTARTS(STR(?property), "http://dbpedia.org/ontology/"))

    \hspace{1cm}FILTER(!CONTAINS(STR(?property), "wikiPage"))

    \hspace{1cm}FILTER(?property NOT IN ([\textit{Values found in Table~\ref{tab:ignored_dbpedia_properties}}]))
    
\}

\vspace{0.5cm}

SELECT ?property ?value WHERE \{

    \hspace{1cm}?value ?property <http://dbpedia.org/resource/\{entity\}> .

    \hspace{1cm}FILTER(STRSTARTS(STR(?property), "http://dbpedia.org/ontology/"))

    \hspace{1cm}FILTER(!CONTAINS(STR(?property), "wikiPage"))

    \hspace{1cm}FILTER(?property NOT IN ([\textit{Values found in Table~\ref{tab:ignored_dbpedia_properties}}]))
    
\}
\end{tcolorbox}

The retrieval of triples contains an initial filtering of NLG-incompatible triples. During this step, triples that contain metadata or wikiPage information were not extracted, (e.g. \texttt{dbo:wikiPageExternalLink}), as well as small list of triples that contained rare Properties that were regarded as irrelevant for NLG (e.g. \texttt{dbo:logo}), and Properties that were frequently linked to list entities (marked by a double underscore (\_\_)), such as those pages that pertaining to a specific title someone had held (for example , \texttt{dbo:politicalLeader}). A full list of these Properties is included in Table~\ref{tab:ignored_dbpedia_properties} in Appendix~\ref{app:rare-props}.

A limit of 50 triples per Property was applied to avoid the creation of datasets heavily skewed by more prolific \texttt{dbo:} Properties such as \texttt{dbo:birthPlace} or \text{dbo:hometown}, which can happen thousands of times with the same Object. In total, we collected 575,871 triples for the Top 10,000 dataset, 120,162 triples for the Top 1,000 dataset, and 5,830 triples for the Random 1,000 dataset; since the Top 1,000 entities are a subset of the Top 10,000 entities, the extracted triples largely overlap,\footnote{Not fully because the queries were made at different times and DBpedia is a living resource in constant evolution.} so the count of unique triples collected is around 600,000. The breakdown of the count by categories is shown in Tables~\ref{tab:val-top1000}, \ref{tab:val-top10000} and \ref{tab:val-random1000}.

\subsection{Validation strategy}
\label{subsec:method-validation}
In the previous subsections, we describe how we select entities and collect triples for each of them; in this section, we describe how each triple is checked and assigned a validation label.

To validate a triple, four sets of values are needed: the expected domain (D$_{Exp}$), or to which class(es) the Property expects its Subject to belong, the expected range (R$_{Exp}$) or which class(es) the Property expects its Object to belong, the actual domain (D$_{Act}$), i.e. the actual class(es) the Subject belongs to, and the actual range (R$_{Act}$), i.e. the actual class(es) the Object belongs to. On DBpedia, the actual values typically take the form of a list, while the expected values are typically atomic or undefined. The classes are encoded under the \texttt{rdf:}Type Property of each entity, using the DBpedia ontology (\texttt{dbo:}). Most types will have a superclass (e.g. \texttt{dbo:Animal} for \texttt{dbo:Person}). In order to get a comprehensive coverage, superclasses are included into the actual domain and actual range lists. We then try and match the expected and actual domain and the expected and actual range, and assign one of four validity labels according to the result:
\begin{itemize}
    \item \textit{Invalid}: An actual type contradicts one expected type;
    \item \textit{Possibly$\_$Valid$_{E}$}: No contradiction, but an entity does not have a dbo:type on DBpedia;
    \item \textit{Possibly$\_$Valid$_{P}$}: No contradiction, but a Property does not specify the type of Subject and/or Object it expects;
    \item \textit{Valid}: Actual and expected types match.
\end{itemize}

\begin{table*}[!bth]
\caption{Categorisation of all Subject and Object type combinations. A, B, C and D are fictitious examples of rdf:Type. \_ symbolises the absence of a defined type. \textbf{D$_{Exp/Act}$}: Expected/Actual Domain; \textbf{R$_{Exp/Act}$}: Expected/Actual Range; \textbf{Possibly$\_$valid$_{E}$}: Undefined Entity type; \textbf{Possibly$\_$valid$_{P}$}: Undefined Property.}
\label{tab:all-combinations}
\centering
\begin{tabular}{|c|c|c|c|p{2.2cm}|p{3cm}|p{3cm}|}
\hline
\textbf{D$_{Exp}$} & \textbf{R$_{Exp}$} & \textbf{D$_{Act}$} & \textbf{R$_{Act}$} & \textbf{Label} & \textbf{Example} & \textbf{Explanation} \\
\hline
\_ & B & Any & D & Invalid & Rutobwe $\|$ country $\|$ Africa & R$_{Exp}$:Country R$_{Act}$:Continent \\
\hline
A & \_ & C & Any & Invalid & East\_Coast\_Road $\|$ map $\|$ Chennai & D$_{Exp}$:Place D$_{Act}$:Road \\
\hline
A & B & C & B & Invalid & Parker\_Dam $\|$ river $\|$ Colorado\_River & D$_{Exp}$:Place D$_{Act}$:Dam \\
\hline
A & B & A & D & Invalid & Nepoko\_River $\|$ mouthMountain $\|$ Africa & R$_{Exp}$:Mountain R$_{Act}$:Continent \\
\hline
A & B & C & D & Invalid & Vudumane\_(singer) $\|$ hometown $\|$ Africa & D$_{Exp}$:Agent D$_{Act}$:Person 

R$_{Exp}$:Settlement R$_{Act}$:Continent \\
\hline
A & B & \_ & D & Invalid & *LooCafe $\|$ foundedBy $\|$ Abhishek\_Nath & R$_{Exp}$:Agent R$_{Act}$:Person \\
\hline
A & B & C & \_ & Invalid & Medieval\_folk\_rock $\|$ instrument $\|$ Singing &  D$_{Exp}$:Artist D$_{Act}$:MusicGenre \\
\hline
\_ & Any & Any & \_ & Possibly$\_$valid$_{E}$& Seven\_Years'\_War $\|$ place $\|$ Americas & D$_{Exp}$:\_ D$_{Act}$:MilitaryConflict

R$_{Exp}$:PopulatedPlace R$_{Act}$:\_ \\
\hline
Any & \_ & \_ & Any & Possibly$\_$valid$_{E}$ & **Tim\_Finnegan $\|$ academicDiscipline $\|$ Chocolate & D$_{Exp}$:AcademicJournal D$_{Act}$:\_

R$_{Exp}$:\_ R$_{Act}$:Food\\
\hline
Any & B & \_ & B & Possibly$\_$valid$_{E}$ & **Loch\_Ness\_Monster $\|$ lake $\|$ Loch\_Ness & D$_{Exp}$:Country D$_{Act}$:\_

R$_{Exp}$:Lake R$_{Act}$:Lake\\
\hline
A & Any & A & \_ & Possibly$\_$valid$_{E}$ & Indiga $\|$ mouthPlace $\|$ Barents\_Sea & D$_{Exp}$:River D$_{Act}$:River

R$_{Exp}$:PopulatedPlace R$_{Act}$:\_\\
\hline
Any & Any & \_ & \_ & Possibly$\_$valid$_{E}$ & Thai\_language $\|$ spokenIn $\|$ http://www4.wiwiss.fu-berlin.de/factbook/resource/Malaysia & D$_{Exp}$:Language D$_{Act}$:\_

R$_{Exp}$:PopulatedPlace R$_{Act}$:\_\\
\hline
\_ & \_ & Any & Any & Possibly$\_$valid$_{P}$ & Beaufort\_Sea $\|$ type $\|$ Sea & D$_{Exp}$:\_ D$_{Act}$:BodyOfWater

R$_{Exp}$:\_ R$_{Act}$:\_ \\
\hline
\_ & B & Any & B & Possibly$\_$valid$_{P}$ & Chukchi\_Sea $\|$ country $\|$ Russia & D$_{Exp}$:\_ D$_{Act}$:Sea

R$_{Exp}$:Country R$_{Act}$:Country\\
\hline
A & \_ & A & Any & Possibly$\_$valid$_{P}$ & North\_Rona $\|$ archipelago $\|$ Atlantic\_Ocean & D$_{Exp}$:Island D$_{Act}$:Island

R$_{Exp}$:\_ R$_{Act}$:Ocean\\
\hline
A & A & B & B & Valid & Zhuang\_Nu $\|$ deathPlace $\|$ Chongqing & D$_{Exp}$:Animal D$_{Act}$:Animal

R$_{Exp}$:Place R$_{Act}$:Place\\
\hline
\end{tabular}
\end{table*}

For instance, a triple such as \texttt{dbr:Ibn\_al-Tilmidh $\|$ dbo:occupation $\|$ dbr:Baghdad} would be marked as \textbf{invalid}, as \texttt{dbo:occupation} expects an entity typed as \texttt{dbo:PersonFunction} as its range, and the actual value is a \texttt{dbo:City}.
A triple such as \texttt{dbr:Caeau\_Ty’n-llwyni $\|$ dbo:areaOfSearch $\|$ dbr:Wales} is marked as \textbf{possibly valid} as \texttt{dbr:Wales} does not have any \texttt{dbo:} class types. The triple \texttt{dbr:Ibn\_al-Tilmidh $\|$ dbo:birthPlace $\|$ dbr:Baghdad} is \textbf{valid}, as \texttt{dbr:Ibn\_al-Tilmidh} is classified as a \texttt{dbo:Person}, \texttt{dbr:Baghdad} is classified as a \texttt{dbo:City}, and \texttt{dbo:birthPlace} expects \texttt{dbo:Animal} as the domain and \texttt{dbo:Place} as the range, which are superclasses of \texttt{dbo:Person} and \texttt{dbo:City} respectively. Table~\ref{tab:all-combinations} explains the different combinations of expected and actual type values, and how they are labelled.\footnote{Not all of these combinations were found in the three datasets. The example taken directly from DBpedia is marked with an asterisk (*), and handcrafted examples are marked with a double asterisk (**).} A local version of entity types, Property definitions and superclasses was created to speed up the validation process used here and during the filtering step in Section~\ref{sec:datasetTriples}.

\subsection{Triple assessment results on three sets of entities}
\label{subsec:method-issues}
Each of the $\sim$600,000 triples for the entities in the three datasets described in Section~\ref{subsec:entity-selection} was assessed as \textit{Invalid}, \textit{Possibly$\_$Valid$_{E}$}, \textit{Possibly$\_$Valid$_{P}$}, or  \textit{Valid} (see Section~\ref{subsec:method-validation}). To get a clearer high-level view of the results, we grouped here \textit{Possibly$\_$Valid$_{E}$} and \textit{Possibly$\_$Valid$_{P}$} in a single category, \textit{Possibly$\_$Valid} (see Section~\ref{subsec:method-diagnosis} for more details). 
For each category, we then calculated the percentage of valid triples, the percentage of non-invalid triples (possibly valid and valid triples together), as well as the ratio of triples and valid triples to entity. Tables~\ref{tab:val-top1000},~\ref{tab:val-top10000} and~\ref{tab:val-random1000} report these numbers along with the number of triples and of entities in each category, for the Top 1,000, Top 10,000 and Random 1,000 entities respectively.

In the Top 1,000 entities shown in Table~\ref{tab:val-top1000}, only a small proportion of the triples have contradictions between expected and actual entity types: the proportion of valid and possibly valid triples ranges from 74$\%$ to 100$\%$ across categories, with an average of 86$\%$. However, when looking a valid triples only, the percentages range from 0$\%$ to 49$\%$, with a very low average of 13$\%$. In other words, in most triples, there are no contradictions between the expected and actual types because the expected and/or the actual type is/are not defined. Across categories, Geography has the highest ratio of valid and possibly valid triples to entity (361). The next three most productive categories were History (86), Food and Agriculture (71), and Arts and recreation (69). In terms of valid triples, two categories stand out, Geography and Biography, with close to 50$\%$ of valid triples; these categories are more likely to provide good data to serve as input for NLG systems.

\begin{table*}[!tbh]
\caption{Validity of triples from the \textbf{Top 1,000} entities.}
\label{tab:val-top1000}
\centering
\begin{tabularx}{\textwidth}{|l|p{1.5cm}|p{1.5cm}|p{1.5cm}|p{1.5cm}|p{1.5cm}|X|}
\hline
\textbf{Category} & \textbf{Number of Entities} & \textbf{Number of triples} & \textbf{Triple to entity ratio} & \textbf{Valid and Possibly valid triple to entity ratio} & \textbf{Valid and Possibly valid triples ($\%$)} & \textbf{Valid triples ($\%$)} \\
\hline
\textit{Arts and recreation} & 75 & 8,207 & 109 & 69 & 84$\%$ & 7$\%$ \\
\hline
\textit{Biography} & 204 & 11,926 & 58 & 50 & 86$\%$ & 46$\%$ \\
\hline
\textit{Food and agriculture} & 34 & 2,618 & 77 & 71 & 96$\%$ & 0$\%$ \\
\hline
\textit{Geography} & 146 & 63,891 & 438 & 361 & 83$\%$ & 49$\%$ \\
\hline
\textit{History} & 46 & 4,929 & 107 & 86 & 87$\%$ & 37$\%$ \\
\hline
\textit{Language and literature} & 46 & 4,483 & 97 & 55 & 81$\%$ & 10$\%$ \\
\hline
\textit{Measurements} & 12 & 171 & 14 & 9 & 100$\%$ & 0$\%$ \\
\hline
\textit{Philosophy} & 13 & 1,039 & 80 & 59 & 79$\%$ & 0$\%$ \\
\hline
\textit{Religion} & 22 & 2,321 & 106 & 67 & 96$\%$ & 0$\%$ \\
\hline
\textit{Science} & 259 & 9,133 & 35 & 25 & 74$\%$ & 5$\%$ \\
\hline
\textit{Social sciences} & 77 & 5,842 & 76 & 45 & 80$\%$ & 4$\%$ \\
\hline
\textit{Technology} & 67 & 5,602 & 84 & 67 & 89$\%$ & 0$\%$ \\
\hline
\textit{\textbf{Average}} & 83 & 10,014 & 107 & 80 & 86$\%$ & 13$\%$ \\
\hline
\end{tabularx}
\end{table*}

Expanding to the top 10,000 entities, shown in Table~\ref{tab:val-top10000}, the general picture is very similar to the Top 1,000 entities, although there are half the average number of triples per entity and half the average number of valid and possibly valid triples per entity. The average proportion of valid and possibly valid triples is 80$\%$ (86$\%$ for the Top 1,000), and the average percentage of valid triples is 10$\%$ (13$\%$ for the Top 1,000). The categories for the Top 10,000 entities are (by design) slightly different from the ones of the Top 1,000, but give a similar image. Geography still has the greatest number of entities, greatest number of triples per entity, and has the highest validity score (51$\%$), and People, which roughly corresponds to Top 1,000's Biography, comes close second with 48$\%$ of valid triples.

\begin{table*}[!tbh]
\caption{Validity of triples from the \textbf{Top 10,000} entities.}
\label{tab:val-top10000}
\centering
\begin{tabularx}{\textwidth}{|X|p{1.5cm}|p{1.5cm}|p{1.5cm}|p{1.5cm}|p{1.5cm}|p{2cm}|}
\hline
\textbf{Category} & \textbf{Number of Entities} & \textbf{Number of triples} & \textbf{Triple to entity ratio} & \textbf{Valid and Possibly valid triple to entity ratio} & \textbf{Valid and Possibly valid triples ($\%$)} & \textbf{Valid triples ($\%$)} \\
\hline
\textit{Anthropology, psychology and everyday life} & 138 & 4,111 & 30 & 16 & 61$\%$ & 5$\%$ \\
\hline
\textit{Arts and recreation} & 646 & 28,874 & 45 & 27 & 85$\%$ & 8$\%$ \\
\hline
\textit{Biology and health sciences} & 1,100 & 18,107 & 16 & 12 & 77$\%$ & 8$\%$ \\
\hline
\textit{Geography} & 1,000 & 266,450 & 266 & 222 & 87$\%$ & 51$\%$ \\
\hline
\textit{History} & 802 & 38,043 & 47 & 39 & 91$\%$ & 40$\%$ \\
\hline
\textit{Language and Literature} & 308 & 14,141 & 46 & 29 & 83$\%$ & 9$\%$ \\
\hline
\textit{Mathematics} & 300 & 2,967 & 10 & 5 & 50$\%$ & 0$\%$ \\
\hline
\textit{People} & 1,943 & 82,538 & 42 & 33 & 79$\%$ & 48$\%$ \\
\hline
\textit{Philosophy} & 101 & 4,524 & 45 & 32 & 86$\%$ & 0$\%$ \\
\hline
\textit{Physical sciences} & 1,317 & 15,170 & 12 & 8 & 74$\%$ & 7$\%$ \\
\hline
\textit{Religion and theology} & 292 & 7,690 & 26 & 17 & 93$\%$ & 3$\%$ \\
\hline
\textit{Society and social sciences} & 1,006 & 52,506 & 52 & 35 & 84$\%$ & 15$\%$ \\
\hline
\textit{Technology} & 1,048 & 40,750 & 39 & 33 & 91$\%$ & 2$\%$ \\
\hline
\textit{\textbf{Average}} & 769 & 44,298 & 52 & 39 & 80$\%$ & 15$\%$ \\
\hline
\end{tabularx}
\end{table*}

\begin{table*}[!tbh]
\caption{Validity of triples  from the \textbf{Random 1,000} entities.}
\label{tab:val-random1000}
\centering
\begin{tabularx}{\textwidth}{|l|p{1.5cm}|p{1.5cm}|p{1.5cm}|p{1.5cm}|p{1.5cm}|X|}
\hline
\textbf{Category} & \textbf{Number of Entities} & \textbf{Number of triples} & \textbf{Triple to entity ratio} & \textbf{Valid and Possibly valid triple to entity ratio} & \textbf{Valid and Possibly valid triples ($\%$)} & \textbf{Valid triples ($\%$)} \\
\hline
\textit{Arts and recreation} & 194 & 992 & 5 & 3 & 64$\%$ & 37$\%$ \\
\hline
\textit{Biography} & 325 & 2,164 & 7 & 5 & 78$\%$ & 54$\%$ \\
\hline
\textit{Food and agriculture} & 2 & 11 & 6 & 5 & 91$\%$ & 27$\%$ \\
\hline
\textit{Geography} & 227 & 1,939 & 9 & 7 & 91$\%$ & 47$\%$ \\
\hline
\textit{History} & 28 & 88 & 3 & 3 & 94$\%$ & 49$\%$ \\
\hline
\textit{Language and literature} & 17 & 58 & 3 & 3 & 97$\%$ & 67$\%$ \\
\hline
\textit{Measurements} & 3 & 1 & 0 & 0 & 100$\%$ & 0$\%$ \\
\hline
\textit{Philosophy} & 1 & 9 & 9 & 8 & 89$\%$ & 0$\%$ \\
\hline
\textit{Religion} & 6 & 11 & 2 & 2 & 100$\%$ & 82$\%$ \\
\hline
\textit{Science} & 89 & 53 & 1 & 1 & 98$\%$ & 64$\%$ \\
\hline
\textit{Social sciences} & 38 & 87 & 2 & 2 & 98$\%$ & 41$\%$ \\
\hline
\textit{Technology} & 69 & 417 & 6 & 5 & 95$\%$ & 44$\%$ \\
\hline
\textit{\textbf{Average}} & 83 & 486 & 4 & 4 & 86$\%$ & 43$\%$ \\
\hline
\end{tabularx}
\end{table*}

For the Random 1,000 entities, which gives a view of what the situation is for most entities on DBpedia, the picture is quite different; see Table~\ref{tab:val-random1000}. While the average number of triples per entity drops to 4 across categories, the overall quality of the triples is higher, with 43$\%$ on average of valid triples, and 86$\%$ of valid and possibly valid triples. Geography and Biography make up more than half of the entities, and together with Arts and Recreation they amount to almost 75$\%$ of all entities. Geography and Biography have more triples per entities on average, but do not have the highest ratios of valid triples, although these ratios are in line with the ones of the Top 1,00 and Top 10,000 entities ($\sim$50$\%$).

\subsection{A preliminary diagnosis and recovery strategy}
\label{subsec:method-diagnosis}

In Section~\ref{subsec:method-issues}, the discrepancy between the last two columns of the three tables shows that only a fraction of triples is explicitly validated, and that a very large number of triples have undefined (missing) expected or actual domain or range. Since we are targeting the collection of valid inputs for NLG systems, we do need validated data. Excluding possibly valid triples would leave us with only a fraction of the data available on DBpedia: for instance, for the 1,048 Technology triple sets in Table~\ref{tab:val-top10000}, 89$\%$ of the triples are possibly valid, and only 2$\%$ are valid. Recovering possibly valid triples, by making them valid, can dramatically shift the numbers and give access to more data on DBpedia.

As described in Section~\ref{subsec:method-validation}, possibly valid triples have one or both of the following issue: a Property definition does not specify which entity class is expected as Subject and/or Object (\textit{Possibly$\_$Valid$_{P}$}), or an entity is not assigned any class on DBpedia (\textit{Possibly$\_$Valid$_{E}$}). As shown in Table ~\ref{tab:four-validities}, a very large majority of the possibly valid triples actually belong to the first type, that is, underspecified Property definitions are a major source of absence of validity. We examined the Properties involved in the labelling of a triple as invalid or possibly valid, so as to isolate individual Properties which could have a high impact on the quality of the collected data.

\begin{table*}[!bth]
\caption{Validity of triples from the three datasets.}
\label{tab:four-validities}
\centering
\begin{tabularx}{\textwidth}{|l|X|X|X|X|}
\hline
\textbf{Dataset} & \textbf{Valid triples} & \textbf{Possibly$\_$Valid$_{P}$} & \textbf{Possibly$\_$Valid$_{E}$} & \textbf{Invalid triples} \\
\hline
\textit{Top 1,000} & 40,147 & 53,444 & 6,923 & 19,648 \\
\hline
\textit{Top 10,000} & 206,543 & 242,052 & 39,992 & 87,284 \\
\hline
\textit{Random 1,000} & 2,796 & 1,571 & 423 & 1,040 \\
\hline
\end{tabularx}
\end{table*}

Looking at Properties responsible for the \textbf{invalid triples}, there is a high level of crossover between the Top 1,000 and the Top 10,000 Properties, as 17 Properties appear in both lists, which is to be expected as the Top 10,000 entity list is an extension of the Top 1,000 one; see Tables~\ref{tab:invalid_prop_t1000} and \ref{tab:invalid_prop_t10000} in Appendix~\ref{app:props-invalid-triples}. A more telling finding is that 8 of the 15 Properties that occur in at least 1$\%$ of the invalid triples from the Random 1,000 dataset shown in Table~\ref{tab:invalid_prop_r1000} (Appendix~\ref{app:props-invalid-triples}) also occur in the Top 1,000 and Top 10,000 datasets: \texttt{team}, \texttt{starring}, \texttt{producer}, \texttt{associatedMusicalArtist}, \texttt{hometown}, \texttt{artist}, \texttt{city} and \texttt{country}. Together, these Properties amount to 19$\%$,  27$\%$ and  62$\%$ of the invalid cases in Top 1,000, Top 10,000 and Random 1,000 datasets respectively. 

A similar effect is found in the to Properties that occur in \textbf{possibly valid triples}: 8 of the 20 Properties in triples for the Random 1,000 entities occur in the Top 1,000 and Top 10,000 tables: \texttt{subdivision}, \texttt{location}, \texttt{country}, \texttt{type}, \texttt{city}, \texttt{education}, \texttt{battle} and \texttt{knownFor}, accounting for 47$\%$, 34$\%$ and 39$\%$ of the possibly valid cases respectively . There is overall an overlap of 14 Properties between the Properties in the Top 1,000 and the Top 10,000 tables; see Tables~\ref{tab:partially_valid_prop_t1000}, \ref{tab:partially_valid_prop_t10000} and \ref{tab:partially_valid_prop_r1000} in Appendix~\ref{app:props-possibly-valid-triples}. 

In other words, defining the expected domain, the expected range, or both for Properties occurring often in invalid and possibly valid triples, will make it possible to validate a large amount of triples that would not be validated otherwise.

\section{Towards high quality inputs for Natural Language Generation}
\label{sec:datasetTriples}
In this section, we describe our approach for optimising both the quantity and the quality of the triples in the triple sets when collecting triples about any entity. 
We first carry out a pilot quality assessment to judge the ability of the validation strategy to correctly identify valid triples, as well as to define human annotation guidelines for the creation of reference data (Section~\ref{subsec:pilotAssessment}). We then show how we apply the validation strategy presented in Section~\ref{subsec:method-validation} to filter out triples and reduce the amount of non-valid triples in our data, and how we use the findings of Section~\ref{subsec:method-diagnosis} to increase the amount of validated triples collected (Section~\ref{subsec:filterAndIncreaseQuantity}). Finally, we report on two evaluations against manually annotated data to show the precision and recall for each step of the triple collection (Section~\ref{subsec:redefineProperties}).

\subsection{Preliminary assessment of validation and evaluation guidelines}
\label{subsec:pilotAssessment}
In order to have an idea about the impact of the basic validation strategy on triple selection, and to establish guidelines for annotating the validity of triples in Section~\ref{subsec:redefineProperties}, we carried out a pilot assessment of the triples of a set of entities. We collected triple sets for 43 entities from the Top 1,000 set (stratified by category), so as to have triples for least one entity per category, following the method described in Section~\ref{subsec:triple-selection}, but allowing for up to 100 instances of the same Property per entity to get a wide range of triples (e.g. in a triple set about Cairo, it is possible to find up to 100 triples \texttt{Subject} $\|$ \texttt{birthPlace} $\|$ \texttt{Cairo}, with 100 different Subject values). This amounted to 3,692 triples in total across the 43 entities. We then ran the validation rules described in Section~\ref{subsec:method-validation} for each triple. For each entity, two lists of triples were stored in a spreadsheet, a list of valid triples (\textit{Valid}, \textit{Possibly$\_$Valid$_{E}$}, \textit{Possibly$\_$Valid$_{P}$}), and a list of invalid triples (\textit{Invalid}). Two annotators A1 and A2 (authors) then annotated each row as ``Correct'' or ``Incorrect'' (about 80 minutes work for each annotator). System precision and recall were computed using each annotator as reference:
A1 - 0.874 precision, 0.866 recall, 0.870 F1; A2 - 0.881 precision, 0.866 recall, 0.873 F1. A1 and A2 then discussed the annotations, for the following outcome:

\begin{itemize}
\item Triple collection (for the study in Section 3.2):
\begin{itemize}
    \item  Triple sets in which the main entity only appears as Object are very likely to contain bad triples.\footnote{These triple sets, even if they were made only of valid triples, are difficult to use in an NLG setting, because the entity is only used to convey information about the Subject entity, which is the focus of the triple. It is challenging to build a narrative around an entity that is never the focus of any triple.} If excluding these triples, for A1 precision and recall reach 0.995 and 0.901, and for A2 0.937 and 0.900 respectively. Triple sets in which the entity never occurs as Subject were thus added to the pre-validation filtering step of Section~\ref{subsec:triple-selection}.
    \item Up to 100 instances of each Property biases the results, because Properties who can reach this count (such as \texttt{birthPlace}, \texttt{birthDate}, \texttt{location}, etc.) tend to be widely valid and correctly labelled so. In order to avoid skewing the results, the final evaluation will limit the number of instance of a Property to 10 for one entity.
    \item The validation approach during triple collection is ready for evaluation.
\end{itemize}
\item Evaluation (for crafting guidelines for the evaluation of the validity of triples in Section~\ref{subsec:redefineProperties}):
\begin{itemize}
    \item  The annotation in terms of Correct/Incorrect was confusing, and there was a risk of bias knowing how the triple had gone through validation. The final evaluations should be done on the raw triples in terms of Valid/Invalid, and Valid and Invalid need to be clearly defined.
    \item  An intermediate category between Valid and Invalid is needed, to give the annotator a chance to label borderline cases.
    \item Possibly valid triples should be considered as invalid to maximise the quality of the triples. 
\end{itemize}
\end{itemize}

\subsection{Validating and recovering triples}
\label{subsec:filterAndIncreaseQuantity}

In this section, we show how the number of collected triples is impacted when applying the validation, and how it is possible to increase the amount of valid triples.

\subsubsection{Impact of the validation on triple filtering}
\label{subsubsec:validation-impact}
We compiled a new stratified sample of 50 entities from the list of \textbf{Top 1,000} to get another representative set across the thirteen categories. We then query the triples for each entity following as described in Section~\ref{subsec:triple-selection}, with the addition of one pre-validation filter and restricting the maximum number of instances of the same Property to 10 for each entity (see Section~\ref{subsec:pilotAssessment}). The the validation rules of Section~\ref{subsec:method-validation} are applied, and invalid triples as well as the two types of possibly valid triples are successively filtered. The successive steps of the process are summarised in Table~\ref{tab:ablation_steps}, which shows the numbers of triples at each numbered filtering/validation step, as well as example triples being filtered.

\begin{table*}[ht]
\caption{The number of triples labelled as \textit{Valid} at each step for the 50 sampled Top 1,000 entities. Steps O$_{1-3}$ denote filters on the original data without validation; steps V$_{1-3}$ denote validation-based filters.}
\label{tab:ablation_steps}
\centering
\begin{tabularx}{\textwidth}{|p{1.25cm}|X|p{1.25cm}|X|}
\hline
\textbf{Filtering step} & \textbf{Description} & \textbf{Number of triples} & \textbf{Example of what gets removed} \\ \hline
O$_0$ & The original dataset & 2,004 & \texttt{Ireland} $\|$ \texttt{wikiPageExternalLink} $\|$ \url{https://books.google.com/books\%3Fid=SJSDj1dDvNUC} \\ \hline
O$_1$ & Triples with list entities (\_\_) removed & 1,895 & \texttt{Philippines $\|$ event $\|$ Philippines\_\_HistoricalEvent\_\_1} \\ \hline
O$_2$ & Triples with specific Properties removed & 1,895 & None \\ \hline
O$_3$  & Triples where the entity from the list only appears as a Object removed & 1,243 & \texttt{Joseph\_Sonnleithner $\|$ knownFor $\|$ Joseph\_Haydn} \\ \hline
V$_1$ & Invalid triples removed & 981 & \texttt{Ibn\_al-Tilmidh $\|$ occupation $\|$ Baghdad} \\ \hline
V$_2$ & Possibly$\_$Valid$_{E}$ triples removed (with undefined entity types) & 954 &\texttt{ Caeau\_Ty'n-llwyni $\|$ areaOfSearch $\|$ Wales} (Wales has no \texttt{dbo:} type) \\ \hline
V$_3$ & Possibly$\_$Valid$_{P}$ removed (with undefined Properties) & 501 & \texttt{Emilian\_dialect $\|$ spokenIn $\|$ Tuscany} \\ \hline
\end{tabularx}
\end{table*}

Table~\ref{tab:ablation_steps} shows that at the end of the validation, only 25$\%$ of the original triples are remaining (501 out of 2,004). After the DBpedia quality assessment and the pilot validation assessment, we know that triples filtered at steps O$_1$ to O$_3$ have a high chance of containing noise and are most likely not good input material for text generation. In other words, we believe that after step O$_3$ (1,243 triples in Table~\ref{tab:ablation_steps}), the concentration of triples that are candidates to be selected as input data for generation is very high, and that the non-valid triples can be recovered.

\subsubsection{Recovering invalid and possibly valid triples}
\label{subsubsec:recovering-triples}

A number of approaches have been developed to automatically correct issues on DBpedia (see Section~\ref{sec:SoA}), but they usually also introduce noise in the data. We note again that after step O$_3$, a large proportion of the triples end up being filtered because of the Property definitions (453/742, 61$\%$). Properties are much easier to fix than entities since there are much less Properties than entities on DBpedia.\footnote{$\sim$1.2K Properties and $\sim$6M entities are currently in use in the English DBpedia.} We also showed in Section~\ref{subsec:method-diagnosis} that only a small set of Properties are responsible for a large proportion of the non-valid triples. By manually fixing Property definitions, it should be possible to make a significant amount of non-valid triples valid; we show in Section~\ref{subsec:redefineProperties} that doing so does increase the final recall of triple selection, without harming the precision.

Examples of \textbf{invalid triples} containing the Properties in Tables~\ref{tab:invalid_prop_t1000}, \ref{tab:invalid_prop_t10000} and \ref{tab:invalid_prop_r1000} (Appendix~\ref{app:rare-props}) were examined in order to provide improved definitions for the Properties. In total there where 35 Properties that accounted for at least 1$\%$ of the problems in one of the datasets. 19 of these were given new definitions, and in the case of the remaining 16 Properties, it was deemed that the issue lay with the entity types more often than not. The new Property definitions are stored in local files, which are checked before the official DBpedia Property definitions during the validation process. In most cases, these changes involved broadening the expected attributes.\footnote{Note that we do not claim to provide a new general-purpose definition for these properties.} For example, the DBpedia definition of \texttt{dbo:academicDiscipline} only accepts entities that are classified as \texttt{dbo:AcademicJournal} as the domain, even though this triple is commonly used for scientists and their area of interest (e.g. \texttt{dbr:Albert\_Einstein} \texttt{dbo:academicDiscipline} 	\texttt{dbr:Physics}). In this instance the definition was changed to also include \texttt{dbo:Scientist} as valid domain values. In triples with \texttt{dbo:mouthPlace}, the opposite was the case as the range was too restricted, and only accepted \texttt{dbo:PopulatedPlace}. This was extended to include any \texttt{dbo:Place}, as triples where the range was a river, sea and ocean were marked as invalid (e.g. \texttt{dbr:Convoy\_PQ\_1} $\|$ \texttt{dbo:place} $\|$ \texttt{dbr:Arctic\_Ocean}). Examples of Properties where the invalidity stemmed from misclassified entities are \texttt{dbr:Mr\_Smith\_\&\_The\_B\_Flat\_Band} $\|$ \texttt{dbo:hometown} $\|$ \texttt{dbr:Europe} or \texttt{dbr:Giorgos\_Mosialos} $\|$ \texttt{dbo:nationality} $\|$ \texttt{dbr:Greek\_language}; these are not fixed by our approach and remain filtered.

A very similar process was applied to the \textbf{possibly valid triples}, however most cases involved the addition of more than one valid domain or range. For example, \texttt{dbo:location} has no expected domain, so five different values were added as valid (\texttt{dbo:Agent}, \texttt{dbo:ArchitecturalStructure}, \texttt{dbo:Event}, \texttt{dbo:Person}, \texttt{dbo:Place}). For \texttt{dbo:product}, four values were added as valid ranges (\texttt{dbo:ArchitecturalStructure}, \texttt{dbo:Beverage}, \texttt{dbo:Food}, \texttt{dbo:Work}). This process was performed for 37 of the 42 Properties in Tables~\ref{tab:partially_valid_prop_t1000}, \ref{tab:partially_valid_prop_t10000} and \ref{tab:partially_valid_prop_r1000} (Appendix~\ref{app:props-possibly-valid-triples}, which are marked in bold.\footnote{ For the remaining Properties, it was quite difficult to define expected domains and/or ranges, for example, \texttt{dbo:knownFor}, \texttt{dbo:religion}, or \texttt{dbo:type}. These were left undefined for the time being.}

Validation checks which included the changes from both the invalid triple and possibly triple analyses were rerun on \textbf{Datasets V$_1$, V$_2$ and V$_3$}, which produced \textbf{Datasets R$_1$, R$_2$ and R$_3$} in Table~\ref{tab:increased_quantity}. The new Property definitions trigger an increase of validated triples of 4$\%$ on V$_1$, 5$\%$ on V$_2$, and up to 55$\%$ on V$_3$, making the count of validated triples go from 501 to 775. In the next section, we provide an evaluation of this approach against a manual annotation of the validity of the triples on two distinct datasets.

\begin{table*}[!tbh]
\caption{The number of triples labelled as \textit{Valid} at each step.}
\label{tab:increased_quantity}
\centering
\begin{tabular}{|c|c|c|c|c|}
\hline
\textbf{Filtering step} & \textbf{Number of triples} & \textbf{Recovery step} & \textbf{Number of triples} & \textbf{Percentage increase} \\
\hline
V$_1$ & 981 & R$_1$ & 1024 & 4$\%$ \\ \hline
V$_2$ & 954 & R$_2$ & 997 & 5$\%$ \\ \hline
V$_3$ & 501 & R$_3$ & 775 & 55$\%$ \\ \hline
\end{tabular}
\end{table*}

\subsection{Evaluation}
\label{subsec:redefineProperties}
In this section, we evaluate the validation and recovery approaches in the context of entity-based triple set selection. For this, we annotated manually the 1,243 triples for 50 entities from the Top 1,000, and another set of 208 triples coming from 50 fully randomly selected entities on DBpedia, to verify if the approach has the potential to be ported to any set of entities.\footnote{Respectively 186 and 81 different unique Properties were found in the two datasets.} Two annotators (authors) annotated each triple in both datasets using the following guidelines, crafted after the pilot evaluation (Section~\ref{subsec:pilotAssessment}):\footnote{A spreadsheet with 30 triples for 1,200 different Properties was also provided to the annotators; the examples were the first 30 triples returned when querying DBpedia for each Property.}

\begin{tcolorbox}[width=\columnwidth, title=Annotation guidelines, label=figure:guidelines for annotation]
1. A triple is annotated as \textbf{Valid} if the knowledge of the triple make sense somehow, independently of how awkward it is to verbalise, the availability of measurements, and of how factually (in)accurate it can be. The triple is clearly correct, and there is no entity or Property substitution that would make it clearly better.

2. A triple is annotated as \textbf{Unsure} if the triple could make sense but it looks like either the Property or the Object is not used exactly as intended. The Property or an Entity was somehow stretched because there is probably no better way to express the knowledge in the triple.

3. A triple is annotated as \textbf{Invalid} when it is clearly wrong, or there is an obvious substitute for an entity or a Property that would make the triple much better.

4. Properties can be checked using the following prefix URL and adding the Property label: \url{https://dbpedia.org/ontology/}, and entities using the following prefix URL and adding the entity label:  \url{https://en.wikipedia.org/wiki/}.
\end{tcolorbox}

For the different parts of the evaluation, we used the following:
\begin{itemize}
    \item Two annotation files with three labels: we use the original annotated files with the three labels --\textit{Valid}, \textit{Unsure} and \textit{Invalid}- for calculating inter-annotator agreement reported in Appendix~\ref{app:eval-details}.
    \item One merged file with two labels: we use a single file with only \textit{Valid} and \textit{Invalid} labels for computing precision, recall and F1 scores. The two annotator annotators discussed and merged their annotation in a single reference file, in which the \textit{Unsure} and \textit{Invalid} labels were grouped under the label \textit{Invalid}. A triple assessed as valid by the validation (before or after recovery) is thus a true positive only if it is marked as valid in the merged file
\end{itemize}

\begin{figure}
    \centering
    \includegraphics[width=1\linewidth]{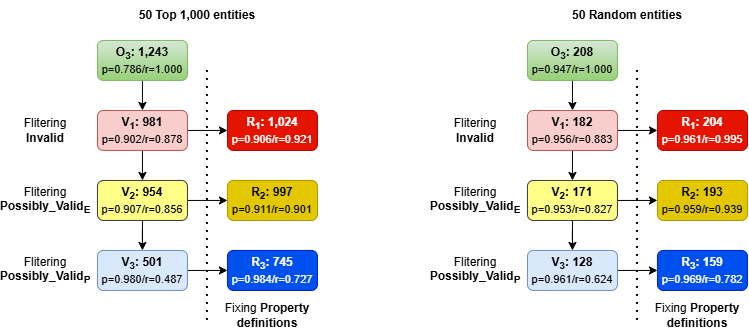}
    \caption{Evaluation results on triples from the 50 random Top 1,000 entities (left) and the 50 fully random entities (right). In each cell: Triple count (top) and precision/recall (bottom) for each combination of filtering and recovery. Precision/recall/F1 numbers on random entities should be taken cautiously given the low amount of triples available for these entities.}
    \label{fig:eval-results}
\end{figure}

Figure~\ref{fig:eval-results} shows the precision and recall scores in both high triple count (Top 1,000 entities) and low triple count (fully random entities) settings. Both settings show similar tendencies: with each validation step, the precision raises while the recall drops. The recovery step substantially increases recall in all cases, with up to 39.5$\%$ increase from V$_3$ (0.487) to R$_3$ (0.727). Tables~\ref{tab:precision_and_recall-top1000} and \ref{tab:precision_and_recall-random} (see Appendix~\ref{app:eval-details}) clearly show that the validation steps reduce the false positives (e.g. from 96 to 10 on the Top 1,000 data), and that the recovery step reduces the false negatives (517 to 275) without adding almost any false positives (10 to 12) for each Validation step. Step R$_3$ achieves a precision of 0.984 and 0.969 on the two datasets respectively, which means that very few of the selected triples are not fully valid. Step R$_1$ achieves the best recall (0.921 and 0.995) and best F1 overall (0.913 and 0.978); this approach can be used if more triples are desired for each entity and triple validation is not the main priority.


\section{Related work}
\label{sec:SoA}
There a body of related work regarding the detection and correction of errors on DBpedia or Linked Open Data in general. Regarding error detection, three main approaches are used: manual~\citep{waitelonis2011whoknows, acosta2018detecting, bu2018using}, semi-automatic~\citep{paulheim2015serving, wu2020guided} and fully automatic~\citep{lehmann2010ore, hao2017cleaning, lertvittayakumjorn2017resolving, caminhas2019detecting}; some experiment with more than one approach~\citep{zaveri2013user}. (Semi-)automatic approaches can use for instance rule crafting~\citep{hao2017cleaning, lertvittayakumjorn2017resolving}, rule learning~\citep{wu2020guided} or different sorts of classifiers~\citep{lehmann2010ore, caminhas2019detecting}. For error correction, most of the works cited above resort to manual correction. Several automatic approaches have been proposed, including via natural language parsing of Wikipedia pages \citep{gangemi2012automatic} or leveraging multilingual versions of DBpedia~\citep{nguyen2016type}. Our work is most similar to that of \citet{paulheim2015serving}, who automatically check statement consistency against the ontology, and then verify the inconsistencies manually before carrying out manual changes.

\section{Conclusions and Future work}
\label{sec:futureWork}
In this paper, we test a validation and recovery approach for the compilation of high precision entity-based triple sets which are intended to be used in the context of multilingual Natural Language Generation. We obtain triple sets in which up to 98$\%$ of the triples are correct according to an evaluation against manually annotated data. Our next logical step is to create challenging input data for NLG systems. One limitation of our approach is that the price to pay for a high precision is a lower recall, so to obtain a variety of large inputs, recovering more triples may be needed. To do so, we plan to test fixing entity types, is in e.g.~\citep{nguyen2016type} or \citep{lertvittayakumjorn2017resolving}.
Using \texttt{dbp:} Properties along with \texttt{dbo:} ones may also reduce the amount of triple sets filtered out. We will also assess the proposed approach on other editions that the English DBpedia (e.g. the Irish Vicip\'eid).

\begin{acknowledgments}
Our work was funded by the Irish Department of Tourism, Culture, Arts, Gaeltacht, Sport and Media via the eST\'OR project, and has also benefitted more generally from being carried out in the wider context of the ADAPT SFI Centre for Digital Media Technology which is funded by Science Foundation Ireland through the SFI Research Centres Programme, and co-funded under the European Regional Development Fund (ERDF) through Grant 13/RC/2106.
\end{acknowledgments}

\section*{Declaration on Generative AI}
 The author(s) have not employed any Generative AI tools for creating the contents of the paper. The authors used Perplexity and Claude 4.8 Opus as support for developing some of the code used  for triple collection and processing. After using this tool, the authors reviewed and edited the content as needed and takes full responsibility for the publication’s content.

\bibliography{custom}

\newpage
\section*{Appendix}
\appendix

\section{Properties not queried}
\label{app:rare-props}
Table~\ref{tab:ignored_dbpedia_properties} shows the list of Properties which are excluded when querying triples for an entity. The list was compiled manually based on observation of the data.

\begin{table*}[ht]
\caption{DBpedia Properties excluded from the SPARQL queries.}
\label{tab:ignored_dbpedia_properties}
\centering
\begin{tabularx}{\textwidth}{|X|X|X|}
\hline
abstract & bicycleInformation & boilerPressure \\ \hline
carNumber & careerStation & collection \\ \hline
damage & depictionDescription & description \\ \hline
event & imageSize & impactFactorAsOf \\ \hline
isHandicappedAccessible & leaderFunction & lengthReference \\ \hline
liberationDate & logo & mapCaption \\ \hline
militaryService & minister & name \\ \hline
note & notes & numberOfVisitorsAsOf \\ \hline
orderInOffice & other & parkingInformation \\ \hline
personFunction & picture & politicalLeader \\ \hline
projectKeyword & pronunciation & quote \\ \hline
reference & restingPlacePosition & restriction \\ \hline
sales & selection & signature \\ \hline
soundRecording & speaker & statisticLabel \\ \hline
strength & termPeriod & thumbnail \\ \hline
title & tournamentRecord & visitorStatisticsAsOf \\ \hline
winsAtAsia & winsAtAus & winsAtChallenges \\ \hline
winsAtChampionships & winsAtJapan & winsAtLET \\ \hline
winsAtNWIDE & winsAtOtherTournaments & winsAtPGA \\ \hline
winsAtSenEuro & winsInEurope & \\ \hline
\end{tabularx}
\end{table*}

\section{Entities used in the pilot assessment study}
\label{app:entities-pilot}
Table~\ref{tab:50_entities} shows random sample entities for the categories used in the Top and Random 1,000 datasets.

\begin{table*}[ht]
\caption{Examples of entities used in the study for each Top and Random 1,000 categories.}
\label{tab:50_entities}
\centering
\begin{tabularx}{\textwidth}{|X|X|}
\hline
\textbf{Category} & \textbf{Entities} \\ \hline
\textit{Arts and recreation} &  Eiffel\_Tower, Trumpet, Music, Martial\_arts \\ \hline
\textit{Biography} &  Martin\_Luther\_King\_Jr., Marco\_Polo, Rembrandt, Johannes\_Brahms, Constantine\_the\_Great, The\_Beatles, Joseph\_Haydn, Napoleon, Ferdinand\_Magellan, Aristotle \\ \hline
\textit{Food and Agriculture} & Agriculture, Spice \\ \hline
\textit{Geography} & Amazon\_River, Oceania, Philippines, United\_States, South\_Pole, Atlantic\_Ocean, Bogotá \\ \hline
\textit{History} & Reformation, American\_Civil\_War \\ \hline
\textit{Language and literature} & Bengali\_language, Grammar \\ \hline
\textit{Measurements} & Second \\ \hline
\textit{Philosophy} & Knowledge \\ \hline
\textit{Religion} & Polytheism \\ \hline
\textit{Science} & Cloud, Mercury, Bacteria, Organism, Camel, Planet, Vacuum, Flower, El\_Niño–Southern\_Oscillation, Force, Symmetry, Algae, Liver, Emotion \\ \hline
\textit{Social sciences} & World\_Health\_Organization, Nationalism, Family \\ \hline
\textit{Technology} & Transistor, Diode, Metallurgy \\ \hline
\end{tabularx}
\end{table*}

\section{Properties in invalid triples}
\label{app:props-invalid-triples}
Tables~\ref{tab:invalid_prop_t1000}, \ref{tab:invalid_prop_t10000} and \ref{tab:invalid_prop_r1000} show the all Properties present in at least 1$\%$ of invalid triples (as detected by the validation rules), ranked by count, for the Top 1,000, the Top 10,000 and the Random 1,000 datasets respectively.

\begin{table*}[ht]
\caption{Common Properties in invalid triples in the \textbf{Top 1,000}. In bold, Properties redefined as part of the recovery step.}
\label{tab:invalid_prop_t1000}
\centering
\begin{tabularx}{\textwidth}{|X|X|X|}
\hline
\textbf{Property} & \textbf{Total} & \textbf{Percentage} \\
\hline
\textbf{academicDiscipline} & 2,759 & 14$\%$ \\
\hline
hometown & 1,579 & 8$\%$ \\
\hline
\textbf{occupation} & 1,368 & 7$\%$ \\
\hline
city & 1,293 & 7$\%$ \\
\hline
\textbf{genre} & 1,067 & 5$\%$ \\
\hline
almaMater & 647 & 3$\%$ \\
\hline
mouthMountain & 640 & 3$\%$ \\
\hline
country & 495 & 3$\%$ \\
\hline
\textbf{place} & 487 & 2$\%$ \\
\hline
\textbf{mouthPlace} & 466 & 2$\%$ \\
\hline
\textbf{locatedInArea} & 447 & 2$\%$ \\
\hline
nationality & 339 & 2$\%$ \\
\hline
\textbf{routeStart} & 308 & 2$\%$ \\
\hline
\textbf{headquarter} & 285 & 1$\%$ \\
\hline
owner & 263 & 1$\%$ \\
\hline
club & 258 & 1$\%$ \\
\hline
\textbf{starring} & 254 & 1$\%$ \\
\hline
award & 238 & 1$\%$ \\
\hline
homeStadium & 226 & 1$\%$ \\
\hline
species & 212 & 1$\%$ \\
\hline
ground & 210 & 1$\%$ \\
\hline
\textbf{producer} & 203 & 1$\%$ \\
\hline
language & 200 & 1$\%$ \\
\hline
Other & 5,404 & 28$\%$ \\
\hline
\end{tabularx}
\end{table*}

\begin{table*}[ht]
\caption{Common Properties in invalid triples in the \textbf{Top 10,000}. In bold, Properties redefined as part of the recovery step.}
\label{tab:invalid_prop_t10000}
\centering
\begin{tabularx}{\textwidth}{|X|X|X|}
\hline
\textbf{Property} & \textbf{Total} & \textbf{Percentage} \\
\hline
\textbf{academicDiscipline} & 11,673 & 13$\%$ \\
\hline
\textbf{occupation} & 6,339 & 7$\%$ \\
\hline
hometown & 5,931 & 7$\%$ \\
\hline
\textbf{starring} & 5,061 & 6$\%$ \\
\hline
city & 3,869 & 4$\%$ \\
\hline
\textbf{producer} & 2,363 & 3$\%$ \\
\hline
\textbf{genre} & 2,231 & 3$\%$ \\
\hline
\textbf{artist} & 2,056 & 2$\%$ \\
\hline
mouthMountain & 1,902 & 2$\%$ \\
\hline
country & 1,808 & 2$\%$ \\
\hline
\textbf{headquarter} & 1,623 & 2$\%$ \\
\hline
almaMater & 1,549 & 2$\%$ \\
\hline
owner & 1,406 & 2$\%$ \\
\hline
\textbf{locatedInArea} & 1,368 & 2$\%$ \\
\hline
nationality & 1,358 & 2$\%$ \\
\hline
associatedMusicalArtist & 1,316 & 2$\%$ \\
\hline
associatedBand & 1,297 & 1$\%$ \\
\hline
award & 1,209 & 1$\%$ \\
\hline
\textbf{mouthPlace} & 1,202 & 1$\%$ \\
\hline
\textbf{routeStart} & 1,160 & 1$\%$ \\
\hline
team & 1,122 & 1$\%$ \\
\hline
\textbf{place} & 1,069 & 1$\%$ \\
\hline
\textbf{territory} & 973 & 1$\%$ \\
\hline
Other & 27,399 & 31$\%$ \\
\hline
\end{tabularx}
\end{table*}

\begin{table*}[ht]
\caption{Common Properties in invalid triples in the \textbf{Random 1,000}. In bold, Properties redefined as part of the recovery step.}
\label{tab:invalid_prop_r1000}
\centering
\begin{tabularx}{\textwidth}{|X|X|X|}
\hline
\textbf{Property} & \textbf{Total} & \textbf{Percentage} \\
\hline
\textbf{team} & 370 & 36$\%$ \\
\hline
\textbf{starring} & 145 & 14$\%$ \\
\hline
currentMember & 67 & 6$\%$ \\
\hline
\textbf{managerClub} & 57 & 5$\%$ \\
\hline
\textbf{producer} & 37 & 4$\%$ \\
\hline
\textbf{routeJunction} & 35 & 3$\%$ \\
\hline
\textbf{associatedMusicalArtist} & 24 & 2$\%$ \\
\hline
\textbf{hometown} & 23 & 2$\%$ \\
\hline
highschool & 16 & 2$\%$ \\
\hline
\textbf{artist} & 15 & 1$\%$ \\
\hline
city & 14 & 1$\%$ \\
\hline
\textbf{firstDriver} & 14 & 1$\%$ \\
\hline
country & 13 & 1$\%$ \\
\hline
birthPlace & 12 & 1$\%$ \\
\hline
\textbf{routeEnd} & 11 & 1$\%$ \\
\hline
Other & 187 & 18$\%$ \\
\hline
\end{tabularx}
\end{table*}

\section{Properties in possibly valid triples}
\label{app:props-possibly-valid-triples}
Tables~\ref{tab:partially_valid_prop_t1000}, \ref{tab:partially_valid_prop_t10000} and \ref{tab:partially_valid_prop_r1000} show the all Properties present in at least 1$\%$ of possibly triples (as detected by the validation rules), ranked by count, for the Top 1,000, the Top 10,000 and the Random 1,000 datasets respectively.

\begin{table*}[ht]
\caption{Common Properties in partially valid triples in the \textbf{Top 1,000}. In bold, Properties redefined as part of the recovery step.}
\label{tab:partially_valid_prop_t1000}
\centering
\begin{tabularx}{\textwidth}{|X|X|X|}
\hline
\textbf{Property} & \textbf{Total} & \textbf{Percentage} \\
\hline
\textbf{location} & 5,009 & 9$\%$ \\
\hline
knownFor & 3,491 & 7$\%$ \\
\hline
\textbf{product} & 3,330 & 6$\%$ \\
\hline
\textbf{industry} & 2,986 & 6$\%$ \\
\hline
type & 2,759 & 5$\%$ \\
\hline
\textbf{country} & 2,637 & 5$\%$ \\
\hline
\textbf{place} & 2,386 & 4$\%$ \\
\hline
\textbf{nonFictionSubject} & 1,979 & 4$\%$ \\
\hline
\textbf{subdivision} & 1,754 & 3$\%$ \\
\hline
\textbf{city} & 1,685 & 3$\%$ \\
\hline
\textbf{citizenship} & 1,428 & 3$\%$ \\
\hline
\textbf{assembly} & 1,331 & 2$\%$ \\
\hline
\textbf{ingredient} & 1,161 & 2$\%$ \\
\hline
\textbf{mainInterest} & 1,054 & 2$\%$ \\
\hline
\textbf{locationCountry} & 1,010 & 2$\%$ \\
\hline
\textbf{language} & 963 & 2$\%$ \\
\hline
\textbf{literaryGenre} & 957 & 2$\%$ \\
\hline
religion & 921 & 2$\%$ \\
\hline
builder & 870 & 2$\%$ \\
\hline
\textbf{influencedBy} & 821 & 2$\%$ \\
\hline
\textbf{education} & 798 & 1$\%$ \\
\hline
\textbf{academicDiscipline} & 755 & 1$\%$ \\
\hline
\textbf{commander} & 743 & 1$\%$ \\
\hline
\textbf{region} & 644 & 1$\%$ \\
\hline
\textbf{service} & 601 & 1$\%$ \\
\hline
\textbf{field} & 577 & 1$\%$ \\
\hline
\textbf{origin} & 536 & 1$\%$ \\
\hline
Other & 10,258 & 19$\%$ \\
\hline
\end{tabularx}
\end{table*}

\begin{table*}[ht]
\caption{Common Properties in partially valid triples in the \textbf{Top 10,000}. In bold, Properties redefined as part of the recovery step.}
\label{tab:partially_valid_prop_t10000}
\centering
\begin{tabularx}{\textwidth}{|X|X|X|}
\hline
\textbf{Property} & \textbf{Total} & \textbf{Percentage} \\
\hline
\textbf{location} & 25,280 & 10$\%$ \\
\hline
type & 15,597 & 6$\%$ \\
\hline
knownFor & 14,632 & 6$\%$ \\
\hline
\textbf{product} & 13,160 & 5$\%$ \\
\hline
\textbf{country} & 12,665 & 5$\%$ \\
\hline
\textbf{industry} & 9,522 & 4$\%$ \\
\hline
\textbf{city} & 8,762 & 4$\%$ \\
\hline
\textbf{place} & 8,497 & 4$\%$ \\
\hline
\textbf{subdivision} & 7,917 & 3$\%$ \\
\hline
\textbf{commander} & 6,290 & 3$\%$ \\
\hline
\textbf{education} & 5,072 & 2$\%$ \\
\hline
\textbf{ingredient} & 4,983 & 2$\%$ \\
\hline
\textbf{nonFictionSubject} & 4,841 & 2$\%$ \\
\hline
\textbf{battle} & 4,625 & 2$\%$ \\
\hline
\textbf{language} & 3,692 & 2$\%$ \\
\hline
\textbf{locationCountry} & 3,288 & 1$\%$ \\
\hline
\textbf{mainInterest} & 3,275 & 1$\%$ \\
\hline
\textbf{citizenship} & 3,231 & 1$\%$ \\
\hline
\textbf{academicDiscipline} & 3,198 & 1$\%$ \\
\hline
\textbf{assembly} & 3,096 & 1$\%$ \\
\hline
religion & 3,069 & 1$\%$ \\
\hline
\textbf{institution} & 3,033 & 1$\%$ \\
\hline
\textbf{service} & 3,002 & 1$\%$ \\
\hline
builder & 2,859 & 1$\%$ \\
\hline
\textbf{literaryGenre} & 2,755 & 1$\%$ \\
\hline
\textbf{region} & 2,679 & 1$\%$ \\
\hline
\textbf{president} & 2,646 & 1$\%$ \\
\hline
Other & 60,386 & 25$\%$ \\
\hline
\end{tabularx}
\end{table*}

\begin{table*}[ht]
\caption{Common Properties in partially valid triples in the \textbf{Random 1,000}. In bold, Properties redefined as part of the recovery step.}
\label{tab:partially_valid_prop_r1000}
\centering
\begin{tabularx}{\textwidth}{|X|X|X|}
\hline
\textbf{Property} & \textbf{Total} & \textbf{Percentage} \\
\hline
\textbf{subdivision} & 267 & 17$\%$ \\
\hline
\textbf{location} & 153 & 10$\%$ \\
\hline
\textbf{country} & 110 & 7$\%$ \\
\hline
type & 86 & 5$\%$ \\
\hline
\textbf{city} & 63 & 4$\%$ \\
\hline
\textbf{occupation} & 57 & 4$\%$ \\
\hline
\textbf{timeZone} & 57 & 4$\%$ \\
\hline
\textbf{predecessor} & 52 & 3$\%$ \\
\hline
position & 51 & 3$\%$ \\
\hline
\textbf{recordLabel} & 50 & 3$\%$ \\
\hline
\textbf{party} & 46 & 3$\%$ \\
\hline
\textbf{successor} & 43 & 3$\%$ \\
\hline
\textbf{genre} & 42 & 3$\%$ \\
\hline
\textbf{distributor} & 35 & 2$\%$ \\
\hline
\textbf{manufacturer} & 33 & 2$\%$ \\
\hline
\textbf{education} & 24 & 2$\%$ \\
\hline
\textbf{battle} & 23 & 1$\%$ \\
\hline
\textbf{album} & 20 & 1$\%$ \\
\hline
\textbf{associatedBand} & 20 & 1$\%$ \\
\hline
knownFor & 19 & 1$\%$ \\
\hline
Other & 320 & 20$\%$ \\
\hline
\end{tabularx}
\end{table*}

\section{Details of the evaluation}
\label{app:eval-details}
Tables~\ref{tab:precision_and_recall-top1000} and \ref{tab:precision_and_recall-random} show the details of the true/false positives/negatives on the 50 random Top 1,000 entities and 50 fully random entities. To have an idea of how challenging it is to follow our annotation guidelines, we calculated inter-annotator agreement on both sets of triples used in the evaluation. Since there are two annotators annotating the same set of items and there are 3 points on the annotation scale, we used Linear weighted Cohen's $\kappa$ \citep{cohen1960kappa}. On the Top 1,000 data, this produced an observed agreement of 0.9 and a $\kappa$ score of 0.75 (substantial agreement \citet{landis1977measurement}). If the Unsure annotations are removed, the $\kappa$ score increases to 0.83, which is interpreted as almost perfect agreement. On the triples from the random entities, the observed agreement is 0.97 and the Linear weighted Cohen's $\kappa$ is 0.70. Note that the numbers of triples collected for the fully random entities are very low, and are only indicative of the appropriateness of the approach on fully random data.

\begin{table*}[!htb]
\caption{Precision and recall statistics at each step (50 randomly selected Top 1,00 entities).}
\label{tab:precision_and_recall-top1000}
\centering
\begin{tabularx}{\textwidth}{|X|X|X|X|X|X|X|X|X|}
\hline
Dataset & True Positives & False Positives & True Negatives & False Negatives & \textbf{Precision} & \textbf{Recall} & F1 \\
\hline
V$_1$ & 885 & 96 & 139 & 123 & 0.902 & 0.878 & 0.890 \\
\hline
V$_2$ & 865 & 89 & 146 & 143 & 0.907 & 0.858 & 0.882 \\
\hline
V$_3$ & 491 & 10 & 225 & 517 & 0.980 & 0.487 & 0.651 \\
\hline
R$_1$ & 928 & 96 & 139 & 80 & 0.906 & \textbf{0.921} & \textbf{0.913} \\
\hline
R$_2$ & 908 & 89 & 146 & 100 & 0.911 & 0.901 & 0.906 \\
\hline
R$_3$ & 733 & 12 & 223 & 275 & \textbf{0.984} & 0.727 & 0.836 \\ 
\hline
\end{tabularx}
\end{table*}

\begin{table*}[!htb]
\caption{Precision and recall statistics at each step (50 fully randomly selected entities).}
\label{tab:precision_and_recall-random}
\centering
\begin{tabularx}{\textwidth}{|X|X|X|X|X|X|X|X|X|}
\hline
Dataset & True Positives & False Positives & True Negatives & False Negatives & \textbf{Precision} & \textbf{Recall} & F1 \\
\hline
V$_1$ & 174 & 8 & 3 & 23 & 0.956 & 0.883 & 0.973 \\
\hline
V$_2$ & 163 & 8 & 3 & 34 & 0.953 & 0.827 & 0.886 \\
\hline
V$_3$ & 123 & 5 & 6 & 74 & 0.961 & 0.624 & 0.757 \\
\hline
R$_1$ & 196 & 8 & 3 & 1 & 0.961 & \textbf{0.995} & \textbf{0.978} \\
\hline
R$_2$ & 185 & 8 & 3 & 12 & 0.959 & 0.939 & 0.949 \\
\hline
R$_3$ & 154 & 5 & 6 & 43 & \textbf{0.969} & 0.782 & 0.865 \\ 
\hline
\end{tabularx}
\end{table*}

\end{document}